\documentclass[letterpaper, 10 pt, conference]{ieeeconf}  

\IEEEoverridecommandlockouts                              

\usepackage{xspace}
\usepackage{colortbl} 
\usepackage{graphicx}
\usepackage{booktabs}
\usepackage{multirow} 
\usepackage[dvipsnames]{xcolor}
\usepackage{caption}
\usepackage{tabularx, microtype}
\usepackage{amsmath}
\usepackage{cuted}
\usepackage{stfloats}
\usepackage{pifont}
\usepackage{amssymb} 
\usepackage{tablefootnote}
\usepackage{float}
\usepackage{makecell}
\usepackage[normalem]{ulem}
\usepackage{bm}
\usepackage{tikz}

\def\eg{\textit{e.g.,}\xspace} 
\def\ie{\textit{i.e.,}\xspace} 
\def\etal{\textit{et al.}\xspace}

\DeclareUnicodeCharacter{21BB}{$\circlearrowleft$}

\newcolumntype{s}{>{\columncolor[gray]{.85}[.5\tabcolsep]}c}
\newcommand{\task}{Interactive VLN in Continuous Environments\xspace}
\newcommand{\taskAcronym}{IVLN-CE\xspace}

\definecolor{lightgray}{RGB}{211,211,211}

\newcommand{\inlineColorbox}[2]{\begingroup\setlength{\fboxsep}{1pt}\colorbox{#1}{\hspace*{2pt}\vphantom{Ay}#2\hspace*{2pt}}\endgroup}

\newcommand\copyrighttext{%
  \footnotesize \textcopyright \the\year{} IEEE. Personal use of this material is permitted. Permission from IEEE must be obtained for all other uses, including reprinting/republishing this material for advertising or promotional purposes, collecting new collected works for resale or redistribution to servers or lists, or reuse of any copyrighted component of this work in other works.}

\newcommand\copyrightnotice{%
\begin{tikzpicture}[remember picture,overlay]
\node[anchor=south,yshift=10pt] at (current page.south) {\fbox{\parbox{\dimexpr0.75\textwidth-\fboxsep-\fboxrule\relax}{\copyrighttext}}};
\end{tikzpicture}%
}

\newcommand{\commonsense}{$^{\star}$\xspace}

\newcommand{\datasetName}{R2RIE-CE\xspace}
\newcommand{\randInter}{Random Interaction}

\newcommand{\numInt}{NI_i}
\newcommand{\numErr}{NE_i}

\newcommand{\interactionLocalization}{I2EDL\xspace}

\newcommand{\instructionTokens}{\Upsilon}

\makeatletter
\let\NAT@parse\undefined
\makeatother

\usepackage{hyperref}

\title{\LARGE \bf
I2EDL: Interactive Instruction Error Detection and Localization
}

\author{Francesco Taioli$^{1,4}$, Stefano Rosa$^{2}$, Alberto Castellini$^{1}$, Lorenzo Natale$^{2}$,\\ Alessio Del Bue$^{2}$, Alessandro Farinelli$^{1}$, Marco Cristani$^{1}$, Yiming Wang$^{3}$
\thanks{$^{1}$ University of Verona, Verona, Italy.}
\thanks{$^{2}$ Istituto Italiano di Tecnologia (IIT), Genova, Italy.}
\thanks{$^{3}$ Fondazione Bruno Kessler, Trento, Italy. }
\thanks{$^{4}$ Polytechnic of Turin, Turin, Italy.\newline \phantom{--------}{\tt\small francesco.taioli@polito.it}}
\thanks{We acknowledge the CINECA award under the ISCRA initiative, for the availability of high-performance computing resources and support. This work was partially sponsored by the PNRR project FAIR - Future AI Research (PE00000013), under the NRRP MUR program funded by the NextGenerationEU. This project has received funding from the European Union's Horizon research and innovation programme G.A. n. 101070227 (CONVINCE).}
}

\begin{document}
\bstctlcite{IEEEexample:BSTcontrol}
\maketitle
\copyrightnotice 
\begin{strip}
  \vspace{-3cm}
  \centering
  \includegraphics[width=1\linewidth]{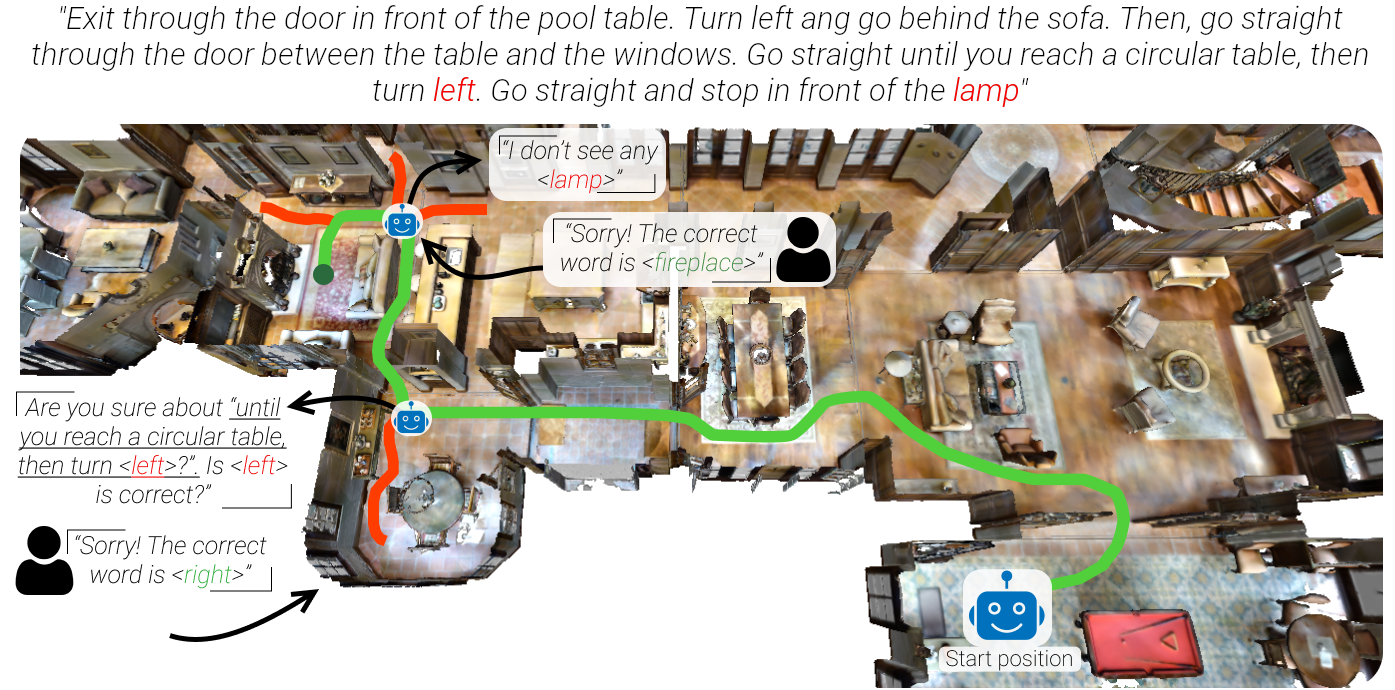}
  \captionof{figure}{A human user guides an agent (bottom right) to reach a target goal ({\color{green}green path}) with instructions expressed in natural language (top) and which may contain errors ({\color{red}red words}). During the navigation, the agent is able to detect and localize instruction errors. Upon detection, the agent asks the user if a particular word in the instruction is correct or not. In case of an incorrect word, the user can reply with the correct one, allowing the agent to resume the navigation. Such human-agent interaction can occur multiple times depending on the error detection algorithm.}
  \label{fig:teaser}
\end{strip}

\thispagestyle{empty}
\pagestyle{empty}

\begin{abstract}
In the Vision-and-Language Navigation in Continuous Environments (VLN-CE) task, the human user guides an autonomous agent to reach a target goal via a series of low-level actions following a textual instruction in natural language. 
However, most existing methods do not address the likely case where users may make mistakes when providing such instruction (\eg ``turn left" instead of ``turn right").
In this work, we address a novel task of \textit{\task~(IVLN-CE)}, which allows the agent to interact with the user during the VLN-CE navigation to verify any doubts regarding the instruction errors. 
We propose an \textit{Interactive Instruction Error Detector and Localizer~(\interactionLocalization)} that triggers the user-agent interaction upon the detection of instruction errors during the navigation. We leverage a pre-trained module to detect instruction errors and pinpoint them in the instruction by cross-referencing the textual input and past observations. In such way, the agent is able to query the user for a timely correction, without demanding the user's cognitive load, as we locate the probable errors to a precise part of the instruction.
We evaluate the proposed \interactionLocalization~on a dataset of instructions containing errors, and further devise a novel metric, the \textit{Success weighted by Interaction Number (SIN)}, to reflect both the navigation performance and the interaction effectiveness.
We show how the proposed method can ask focused requests for corrections to the user, which in turn increases the navigation success, while minimizing the interactions.
\end{abstract}

\section{Introduction}
\label{sec:intro}
The research of Vision-and-Language Navigation (VLN) aims to develop agents that can navigate to a specified location within a 3D space by following  instructions expressed in natural language. This research task aligns with the broader ambition of embodied AI, which allows automated agents to engage with human users via natural language conversations, understanding their surroundings~\cite{Taioli_2023_ICCV}, and executing tasks in the real world. 
There are many benchmark datasets in the literature of VLN, including the seminal dataset Room-to-Room (R2R) for discrete environments operating on discrete navigation graphs~\cite{Anderson_2018_vln}, and the more realistic R2R for continuous environments (R2R-CE) operating via a set of low-level actions to any unobstructed point in a scene~\cite{krantz_vlnce_2020}. Most of the benchmark datasets consider only \emph{correct} language instructions, implying that the users never make mistakes. However, this is not always the case, since instructions can be very complex and the ability of giving right directions vary greatly among people~\cite{lloyd1989cognitive}.
In fact, it is a matter of \emph{spatial cognition} skills, i.e., how humans mentally represent and process spatial information~\cite{liben2022conceptual}, and the ability of creating \emph{cognitive maps}, which are mental representation of the layout and contents of an environment~\cite{couclelis1996verbal}. 

Recently, Taioli \etal~propose the R2RIE-CE benchmark dataset~\cite{iedl}, which introduces wrong instructions in the form of incorrect directions (left, right, etc.) and misplaced rooms or objects, to model the effect of inaccurate cognitive maps. This is helpful in benchmarking the robustness of state-of-the-art VLN policies, that have been designed to cope with ideal interactions only. The R2RIE-CE dataset can also be used to evaluate algorithms for detecting and localizing errors in a given instruction. Yet, their detection and localization baselines operate in an offline mode, \ie the errors in the instructions are individuated only after an agent has finished its search, thus leaving no chance for the agent and the user to interact and recover the errors \emph{while exploring the environment}. 
On the contrary, enabling human-agent interaction during navigation could be effective. As shown in Fig.~\ref{fig:teaser}, a human user may initially give an erroneous instruction, with a wrong direction word \textit{``left"} to \textit{``right"} and a wrong object landmark \textit{``lamp"} to \textit{``fireplace"}, which can cause the agent to deviate from its target position, failing the navigation task. However, if the agent can detect and locate potential errors while navigating and observing the scene, it can prompt the user for instruction corrections, thus improving the success rate of the task. 
Such interactive VLN with error awareness introduces additional challenges on top of the existing VLN task (\eg visual perception, spatial reasoning and vision-language alignment). In particular, the agent should identify potential errors promptly at an early stage, with only partial observations of the scene. 
Moreover, since it is not ideal to have an agent that constantly interacts with a human user asking for potential errors (both for human disturbance and cognitive load), it is essential to have an accurate online instruction error detector and localizer, thus \textit{``asking the right question at the right time."}

In this work, we address \textit{\task} (\textit{\taskAcronym}). Human users are allowed to make errors in their initial instructions and subsequently correct them if the agent accurately detects and locates the errors through human-agent interactions. We propose an effective baseline, named Interactive Instruction Error Detector and Localizer (\textit{\interactionLocalization}), that operates in an online mode given only partial observations. Inspired by the offline method IEDL proposed in~\cite{iedl}, we first collect a set of visual observations from the agent. By leveraging the pre-trained models in IEDL, we can identify errors within the instruction and precisely locate them. Then, upon positive detection, the agent asks the user if a specific word is wrong and, if this is the case, it obtains an accurate replacement.
In addition to common VLN performance metrics, \ie Success Rate (\texttt{SR}) and Success
weighted by Path Length (\texttt{SPL}), we also propose a novel metric that is specific for the \textit{\taskAcronym} task, the \textit{Success weighted by Interaction Number}~(\texttt{SIN}), which reflects both the navigation performance and interaction effectiveness, by encouraging a higher success rate while limiting the interaction numbers.
We evaluate our method on R2RIE-CE under different instruction errors and prove that our baseline is more effective than an agent that randomly interacts with the user. 
In summary, our contributions are listed below:
\begin{itemize}
    \item We establish the \textit{\taskAcronym} task, \ie \textit{\task}, simulating the real world cases where humans are allowed to make mistakes when providing instruction, and agents are allowed to interact with humans to correct them.
    \item We propose an effective baseline, \interactionLocalization, which interacts with the user in an online manner upon detecting instruction errors and prompting the focused question with localized errors.
    
    \item We propose a novel metric that measure the interaction effectiveness in terms of navigation performance by combining the Success Rate and interaction numbers with the user. Our metric serves as a primary quantifier for comparing the performance of different agents. 
\end{itemize}

\section{Related Works}
\label{sec:related}
\textbf{Vision and Language Navigation.}
The task of \emph{Vision-and-Language Navigation} (VLN) was initially introduced in \cite{Anderson_2018_vln}. Early iterations were based on the Matterport3D Sim \cite{matterport_dataset} and on the Room-to-Room (R2R) dataset \cite{Anderson_2018_vln}, which represents the environment as a sparse undirected graph of poses with associated panoramic views. In this discrete environment, agents can only move between pre-existing nodes of the graph.
A continuous environment variant of the task (VLN-CE), together with a new dataset (R2R-CE) was introduced in~\cite{krantz_vlnce_2020}, using the Habitat simulator~\cite{habitat_19_iccv}.
In VLN-CE, agents are required to follow the instructions while navigating freely in the environment rather than teleporting between nodes.
While initial approaches used recurrent representations to encode the agent's history and predict the next action~\cite{hamt}, 
more recent methods proposed to predict candidate waypoints and select the next best goal~\cite{discrete_to_cont}, the use of topological memory~\cite{an2023etpnav} to better encode the history of observations, and metric map pre-training~\cite{an2023bevbert} to increase spatial awareness. 
Recent works have studied how VLN agents use directions and objects to navigate by masking words~\cite{zhu2022diagnosing,limitation_vln_agent_aamas} as well as the performance drop in unseen environments~\cite{diagnosin_env_bias}.
However, current methods and datasets do not consider the case of potentially wrong instructions. We propose to specifically study the case of instructions containing errors.

\textbf{Cooperative Vision-dialog Navigation.} 
The task of \emph{Navigation from Dialogue History} (NDH) was introduced in \cite{cvdn}. In NDH, an agent is given a target object and a dialogue history between two humans cooperating to find it. These approaches are usually evaluated only in terms of progress towards the goal. 
\cite{roman2020rmm} propose turn-based dialogue between two agents: the navigating agent and the guiding agent. Both agents learn to simulate questions or answers by the other.
Vision-dialog navigation has been extended to the real world. \cite{banerjee2021robotslang} proposed RobotSlang, a dataset of natural language dialogues between a human operator teleoperating a robot and a human commander that provides guidance towards a goal.
In \cite{shrivastava2021visitron} the agent is trained to identify when to engage in dialogue with a navigator agent via masking and directional grounding.
\cite{zhu2021self} proposes to factor out the action of querying into two different whether-to-ask and a what-to-ask policies.

\textbf{Interactive VLN.}
More closely related to our proposal are approaches where the agent interacts with an oracle by asking for help.
In \cite{just_ask} the agent uses a dedicated policy action for asking for help. The action of querying the oracle is based on model confusion (\ie the agent is unsure about which action to take next) and is penalized via a negative reward. When queried, the oracle returns the next shortest path action to the goal.
A metric is also introduced to evaluate the effectiveness of human-agent interaction, as the percentage of total ask actions per episode. \cite{just_ask} injects a probability of the oracle making a mistake, to simulate a more realistic user. We directly start with a dataset containing mistakes, to simulate wrong instructions given to the agent.
A different way to limit interactions with the oracle is fixing a query budget. In \cite{vision_based_nav} the agent asks for help when unsure about the next action or lost. Upon being called, the oracle provides a short-term goal in natural language. Intervention could be direct (the oracle takes control of the agent) or indirect (the oracle adds new information via short-term textual instructions). A dedicated policy is trained to ask for intervention based on the budget.
\cite{avlen} address the task of \emph{Audio-Visual-Language Embodied Navigation} (AVLEN). The agent can query an oracle under a budget (\ie the maximum number of queries is limited). The effect of number of interactions on success is not directly evaluated, but only indirectly through success metrics.
\cite{help_anna} relaxes assumptions on the oracle by simulating assistants that are only aware of the agent (and can thus provide assistance) when it enters their zone of attention. Interaction is evaluated in terms of number of requests per task.
In both \cite{avlen} and \cite{help_anna}, the oracle replies in the form of a full textual instruction. In contrast, in our proposal, the oracle only substitutes a wrong word with the correct one in the original instruction.

\section{Our Contributions}
\label{sec:method}
\textbf{Task Formulation of IVLN-CE.} For each episode $i$, a human describes to an agent how to reach a target goal by means of a natural language instruction $\mathcal{I}_i$ composed of $F$ words, \ie $\mathcal{I}_i~=~\{w_1, ..., w_F\}$. Note that the instruction given by the human may contain errors, formally, instruction $\mathcal{I}_i$ can have up to $E$ words that are incorrect.
We then define the instruction embedding as~$\Upsilon_i\in \mathbb{R}^{W \times D}$, \ie the instruction $\mathcal{I}_i$ is tokenized and padded up to $W=80$ tokens, while $D$ is the dimension of the latent space in which each token is projected, following~\cite{an2023etpnav,an2023bevbert}. Without loss of generality, we assume that each word is tokenized into one token.

At each time step $t$, the agent receives a visual observation $O_t$, namely an RGB-D image. Let $T$ be the total number of steps executed by policy $\pi$, we define the set of visual observation $\mathcal{O} = \{O_1, ..., O_T \}$. Policy $\pi$, for every step $t$,  predicts an action $a_t$ in the set $\{$\texttt{Forward 0.25m, Turn Left 15°, Turn Right 15°, Stop}$\}$.

For every episode $i$, and at every step $t$, the agent has the possibility to query the human, checking if a particular token $\ell_i^j$ is correct or not, where $j \in [0, len(\instructionTokens_i) -1]$  and $len(\cdot)$ returns the total number of tokens for
instruction $\instructionTokens_i$.

Asking just a token to the user would be ineffective, since the user would hardly understand the sense of a single word (token), and, in the case of multiple instances of the same word, misunderstandings could easily arise. Therefore, the agent passes to the users a portion (\textit{context}) of the instruction, made by multiple tokens $L_{j,\varsigma_l}=[\ell_i^{j-\varsigma_l}, \ell_i^{j+\varsigma_l}]$, where $\varsigma_l$ is the \emph{contextualization length}. The human, upon receiving a request from the agent, returns the real token if a wrong token is found within $[\ell_i^{j-\tau_l}, \ell_i^{j+\tau_l}]$, where $\tau_l$  is a \textit{localization threshold}. This correction mechanism ensures that the human can provide the correct token even if there is a slight discrepancy of $\tau_l$ tokens in the location pointed out by the agent.

\textbf{Policy.} The agent's policy $\pi$ is implemented in this paper by the current state-of-the-art method for VLN-CE\footnote{\href{https://eval.ai/web/challenges/challenge-page/719/leaderboard/1966/success}{Eval AI - VLN-CE Challenge}}, \ie BEVBert~\cite{an2023bevbert}. However, our approach is model-agnostic. 
BEVBert is composed of three essential modules that allow the agent to balance the demand for short-term reasoning and long-term planning: \textit{(i)}~a graph-based topological map for long-term planning equipped with a global action space; \textit{(ii)}~a local metric map for short-term reasoning equipped with a local action space; \textit{(iii)}~a map-based pre-training paradigm to enhance the spatial-aware cross-modal reasoning. Formally, given an episode $i$, let $t$ be the current time step and $n$ the current node of the topological graph inside the environment. Policy $\pi$ selects the best candidate node from the topological map of point \textit{(i)}, and low-level actions are performed to bring the agent from node $n_t$ to $n_{t+1}$. To be comparable to other VLN-CE agents, we maintain a maximum number of $k=15$ steps, as done in BEVBert~\cite{an2023bevbert}. 
Notice that we do not train the policy $\pi$, as it is considered as given. 
The focus of our work is user-agent interactions during navigation when any instruction error is detected and localized in an online manner.

\textbf{Instruction Error Detection \& Localization.} We employ the recently proposed Instruction Error Detector \& Localizer (\textit{IEDL}) \cite{iedl}. IEDL is firstly composed of a cross-modal transformer, which fuses together the semantic meaning of the language instruction with the sequence of the visual observations of the agent, producing visual-language-aligned features. Then, 
these features are fed to two heads:~\textit{(i)}~a detection head $f_d$, trained to detect when the instruction does not align with the sequence of observations, which outputs an alignment score $a \in [0,1]$;
\textit{(ii)}~ a localization head $f_l$, which predicts the locations of the words that may introduce errors within the instructions. 
Notably, \textit{IEDL} is trained and evaluated in an offline manner where errors are detected and localized after each complete episode. 

\textbf{\interactionLocalization: Interactive-IEDL.} For each episode $i$, we execute the policy $\pi$ following instruction $\mathcal{I}_i$ for at least $p$ steps to acquire  the set of visual observations $\mathcal{O} = \{O_1, ..., O_p \}$.  
When the current step $t \ge p$, we use the detection head $f_d$ of IEDL to check if the alignment score is $a \ge \tau_d$, where $\tau_d = 0.6$ is a \textit{detection threshold}. If the detection is positive, meaning that the instruction contains at least one error, we use the IEDL localization head $f_l$ to localize the errors. Formally, we apply the $softmax$ operator over the output of $f_l$, and then select the token $\ell_i^{j}$ with the highest probability, where $j$ is the index of the token.
When a positive detection occurs, we increment variable $\numInt$, showing that the agent has detected and localized errors in the instruction.
We then simulate an agent-human interaction by asking the question \textit{``I think there is an error in this part of the instruction: $<$part$>$, and specifically on this $<$token$>$. Is this the case?"}
In this question, $<$\textit{token}$>$ refers to token $\ell_i^{j}$, while
$<$\textit{part}$>$ refers to context $L_{j,\varsigma_l}$.
If the range identified by tokens $[\ell_i^{j-\tau_l}, \ell_i^{j+\tau_l}]$ contains the error, then the agent receives the correct token, and the embedding for instruction $\Upsilon_i$ is recomputed.
Otherwise, if the detection is a false positive or no error is found by the human within the specified range, no action is performed by the human.
After the interaction, the agent resumes its navigation, having the possibility to query the user for every step until $T$ steps are performed or action \texttt{Stop} action is selected by the policy, meaning that the agent believes to have reached the goal.

\textbf{Success weighted by Interaction Number.}
Since our interaction scheme for VLN is new, we need to propose a novel figure of merit. The rationale is that we want to weight the success rate, dependently on how many times the agent requires the human intervention: the higher the number of interventions, the less valuable the success rate. We thus propose                 
\texttt{SIN}, \ie \textit{Success weighted by Interaction Number}, specifically designed to combine, in a single measure, both the \texttt{SR} and the number of interactions with the user. 
Inspired by the Success weighted by Path Length metric~\cite{anderson2018evaluation}, we define \texttt{SIN} as:
\begin{equation}
SIN = \frac{1}{N}\sum_{i=1}^{N} S_i \frac{1}{1 + \lambda \frac{\numInt}{max(1, \numErr)}}
\end{equation}
where $\numInt$ is the number of interactions with the user, $\numErr$ is the number of errors in episode $i$ and $S_i$ is a binary indicator of success for episode $i$. 
The $max(\cdot)$ operator at the denominator ensures the number of interactions $\numInt$ are weighted by the number of errors $\numErr$. Note that if no errors are present for episode $i$, $max(1, \numErr)=1$. $\lambda$ is a weighting factor that modulates the penalisation for the interaction numbers.
We consider \texttt{SIN} as the primary metric in evaluating methods addressing the \textit{IVLN-CE} task.

\textbf{SIN properties.} \texttt{SIN} is ranged between 0 and 1. A higher value indicates a better navigation performance with interaction efficiency. Moreover,
the proposed \texttt{SIN} metric possesses several favourable properties:

\textit{(i)} when no interaction is performed with the human (\ie when $\numInt$ is $0$), \texttt{SIN} is mathematically equivalent to \texttt{SR}.  

\textit{(ii)} \texttt{SIN} penalizes false positives detections.

\textit{Proof:} for every correct episode $i$ (\ie the instruction is correct and thus $\numErr = 0$), a perfect agent will not interact with the human, thus $\numInt=0$. If this is not the case, $\numInt$ is increased accordingly, thus minimizing the \texttt{SIN} metric. Note that, in this scenario, the denominator is $1$, resulting in increased importance assigned to each unnecessary interaction.

\textit{(iii)} \texttt{SIN} penalizes repetitive interactions.

\textit{Proof:} for every incorrect episode $i$ (\ie the instructions contains error), the \texttt{SIN} metric will be penalized as the agent request multiple interactions with the human.

\textit{(iv)} the weighting factor $\lambda$ prevents the denominator from becoming excessively large. The metric can still show the improvement in \texttt{SR} while penalizing excessive interaction. 
We found that a $\lambda=0.01$ is a good compromise between weighting \texttt{SR} and number of interactions.

\section{Experiments}
\label{sec:exps}

\begin{table*}[ht!]
\caption{Results show the increase of \texttt{SIN} (in $\%$) under different paradigms of interaction on \datasetName benchmark, with localization threshold $\tau_l=1$, weighting factor $\lambda=0.01$ from step $p=4$ onwards. The primary metric \inlineColorbox{lightgray}{SIN} is highlighted. Under the \emph{``No Interaction"} column, we report the \texttt{SR}, \texttt{SPL} metrics of the BEVBert policy\cite{an2023bevbert}, also showing the Success Rate Upper Bound ($\overline{\texttt{SR}}$). For \textit{\interactionLocalization}, we set detection threshold $\tau_d=0.6$. Error type based on R2RIE-CE Val Unseen Dataset.}
\centering
\resizebox{1\textwidth}{!}{
\begin{tabular}{c cccc scccc scccc scccc }
\midrule
\multirow{2}{*}{Error type} & \multicolumn{2}{m{2.5cm}} {No interaction}&&& \multicolumn{2}{m{2.5cm}}{Random Interaction} &&&& \multicolumn{2}{m{2.5cm}}{Always Ask} &&&& \multicolumn{2}{m{2.5cm}}{\interactionLocalization} &&&  \\ \cmidrule{2-4} \cmidrule{6-9}  \cmidrule{11-14}  \cmidrule{16-19}
                                  
                                  &\scriptsize\textbf{SR}~$\uparrow$ & \scriptsize\textbf{SPL}~$\uparrow$ & \scriptsize\textbf{$\overline{\texttt{SR}}$}~$\uparrow$ && \scriptsize\textbf{SIN}~$\uparrow$ & \scriptsize\textbf{MIN}~$\downarrow$ & \scriptsize\textbf{SR}~$\uparrow$ & \scriptsize\textbf{SPL}~$\uparrow$  && \scriptsize\textbf{SIN}~$\uparrow$ & \scriptsize\textbf{MIN}~$\downarrow$ & \scriptsize\textbf{SR}~$\uparrow$ & \scriptsize\textbf{SPL}~$\uparrow$ && \scriptsize\textbf{SIN}~$\uparrow$ & \scriptsize\textbf{MIN}~$\downarrow$ & \scriptsize\textbf{SR}~$\uparrow$ & \scriptsize\textbf{SPL}~$\uparrow$ &\\ \midrule
                               Direction                  &53.4   &43.5   &58.5    &&52.9  &1.82   &53.6   & 43.6   &&52.8&3.64 &54.3&44.0   &&\textbf{53.2}  & 0.50  &53.4   &43.5  &   \\
                              Room\commonsense            &58.1   &48.6   &60.4    &&57.1  &1.81   &57.9   &48.4                      &&56.8&3.62 &58.4&48.9   &&\textbf{58.1}    & 0.79  &58.4   &48.8  &   \\
                              Object\commonsense          &56.1   &46.1   &58.7    &&55.8  &1.75   &56.6   &46.4          &&55.2&3.53 &56.7&46.7   &&\textbf{56.1}  & 0.70  &56.3   &46.3  &   \\
                              Room\&Object \commonsense   &57.3   &46.9   &61.1    &&56.9  &1.86   &57.4   &47.3                      &&57.3&3.75 &58.4&47.8   &&\textbf{58.3}  &1.15  &58.5   &47.7  &   \\
                              All  \commonsense           &52.4   &42.6   &61.9    &&53.3  &1.97   &53.8   &43.5           &&53.2&3.95 &54.1&43.8   &&\textbf{53.8}  & 1.37  &54.0   &43.4  &   \\ \cmidrule{2-19} 
                              Avg.                        &55.4   &45.5   &60.1    &&55.2  &1.84   &55.9   &45.8                      &&55.0&3.70 &56.4&46.2   &&\textbf{55.9}  & 0.90  &56.1   &46.0      \\ \midrule
\end{tabular}}
\label{table:results_sr_different_method}
\end{table*}

\textbf{Dataset.}
We evaluate our method I2EDL in the recently proposed \datasetName dataset~\cite{iedl}, \ie R2R with Instruction Errors in Continuous Environment. \datasetName is composed of five sets by incorporating various types of instruction errors, including: 
\textit{(i)}~\textit{Direction} (one error),
\textit{(ii)}~\textit{Object} (one error),
\textit{(iii)}~\textit{Room} (one error),
\textit{(iv)}~\textit{Room\&Object }(two errors); and finally
\textit{(v)}~\textit{All} (three errors). 
Each set $\mathcal{E}$ is composed of correct episodes $\mathcal{E}_c$ and perturbed episodes $\mathcal{E}_p$. For each episode $e_i \in \mathcal{E}_c$, the authors derived an associated perturbed episode containing specific instruction errors, which is stored in the associated set $\mathcal{E}_p$.
The ratio of $\mathcal{E}_c$ and $\mathcal{E}_p$ is $50\%$.

\textbf{Metrics.}
To appreciate the qualities of our proposed metric~\texttt{SIN} \textit{w.r.t}. other, standard, VLN metrics~\cite{Anderson_2018_vln,anderson2018evaluation}, we consider:
\textit{(i)} Success Rate (\texttt{SR}): an episode is considered successful if the distance between the final position of the agent and the target location is less than 3 meters. 
\textit{(ii)} Success weighted by Path Length~\cite{anderson2018evaluation} (\texttt{SPL}): defined as 
\[SPL = \frac{1}{N} \sum_{i=1}^{N} S_i \frac{\ell_i}{max(p_i, \ell_i)}\]
where $N$ are the total number of episodes $\mathcal{E}$, $\ell_i$ is the shortest path distance from the agent’s starting position to the goal
in episode $i$, and $p_i$ is the length of the path actually taken by the agent in episode $i$ and $S_i$ is the binary indicator of success for episode $i$. 
Finally, \textit{(iii)} we report \textit{Mean Interaction Number} (\texttt{MIN}), which is defined as 
\[MIN =\frac{1}{N}\sum_{i=1}^{N} \numInt \]
where $\numInt$ is the number of times the agent interacts with the user in episode $i$.

\textbf{Baseline.}
As \textit{\taskAcronym} is a novel task, there exists no baselines. We thus compare our method I2EDL with a \textit{``Random Interaction"} and an \textit{``Always Ask"} baseline.
Specifically, the \textit{``Random Interaction"} for every episode $i$ and for every step, randomly predicts if instruction $\instructionTokens_i$ contains errors. If the detection is positive (\ie the instruction contains an error), we then randomly predict a token $\ell_i^j$, where $j = rand(0, len(\instructionTokens_i) -1)$ and $rand$ returns a random number between the arguments. The \textit{``Always Ask"} baseline prompts an interaction at every step from step $p = 4$ onwards. At each interaction, it randomly predicts the erroneous token in the same way as the \textit{``Random Interaction"} baseline.

\textbf{Success Rate and interaction paradigm}.
In Tab.~\ref{table:results_sr_different_method}, under the \textit{``No interaction"} column, we report the \texttt{SR} and \texttt{SPL} metrics of the current state-of-the-art method~\cite{an2023bevbert} for VLN-CE, operating under the different benchmarks of \datasetName~\cite{iedl}, thus establishing the lower bound of performances of the agent without interaction.
In the \textit{\taskAcronym} scenario, agents should minimize the number of interactions with the human while maximizing their effectiveness.
Thus, in Tab.~\ref{table:results_sr_different_method} we report under the \textit{``Random Interaction"} column the \textit{Success weighted by Interaction Number} (\texttt{SIN}), \textit{Mean Interaction Number} (\texttt{MIN}), \texttt{SR} and \texttt{SPL} scores.
We can see that the \textit{``\randInter"}, without being able to distinguish correct instruction from instruction with errors, performs an average of $\sim2$ interactions per episode, thus annoying users with unnecessary requests. This behaviour is also reflected in the \texttt{SIN} metric, which penalizes unnecessary interactions.
Such behaviour is even more evident with the \textit{``Always Ask"} baseline. By constantly interacting with the user, the \texttt{SR} is high, however, achieved by an annoyingly large number of user-agent interactions.
Finally, in the last columns of Tab.~\ref{table:results_sr_different_method} we present our method \textit{\interactionLocalization}, checking instruction errors from step $p=4$ onwards, with a detection threshold $\tau_d=0.6$ and  
a localization threshold $\tau_l=1$ (\ie the predicted token position should differ at most for 1 token). 
First of all, we note that I2EDL has a much higher \texttt{SIN} than the \textit{``\randInter"}, meaning that our method is able to detect instruction errors and localize them more precisely, thus maximizing the effectiveness of the interactions. This is also reflected under the \texttt{SR} column, in which, apart from the \textit{Direction} error benchmark, \textit{\interactionLocalization} has an equal or better \texttt{SR} performance, while halving the average \texttt{MIN} ($\textbf{0.90}$ \textit{vs} $1.84$) and scoring consistently lower in terms of \texttt{SIN}. Compared to the\textit{ ``Always Ask"} baseline, \textit{I2EDL} has a higher \texttt{SIN} ({$\textbf{55.9}$} \textit{vs} $55.0$) while having an extremely low \texttt{MIN} score of $\textbf{0.90}$ \textit{vs} $3.70$.
Notably, as also reported by~\cite{iedl}, \textit{I2EDL} has the lowest results on the \textit{Direction} error benchmark, indicating the challenging of \datasetName. 

\textbf{What is the SR upper bound on \datasetName?} 
In this experiment, we want to establish the Success Rate upper bound ($\overline{\texttt{SR}}$) that agents can reach in \datasetName, \ie simulating a perfect agent that is not affected by instruction perturbations. Indeed, ideal agents, as humans, are capable of identifying instruction errors, reasoning and automatically recovering from them. To do this, for each perturbed episode $i$ in $\mathcal{E}_p$, we substitute the associated perturbed instruction with the correct one, thus giving the correct instruction from the beginning of the episode. Note that we do not change instructions for correct episode $i$ in $\mathcal{E}_c$. We report the results in Tab.~\ref{table:results_sr_different_method}, under the $\overline{\texttt{SR}}$ metric. As we can see, the biggest increment is in the \textit{All} benchmark ($52.4$ vs $61.9$), in which the wrong episode $i \in \mathcal{E}_p$ contains three errors. 
Notably, the second biggest increment is in the \textit{Direction} benchmark,
which brings the \texttt{SR} from $53.4$ to $58.5$. Overall, on average, we have an $8.48\%$ increment from $55.4$ to $60.1$.

\begin{figure}[h!]
    \includegraphics[width=1\linewidth]{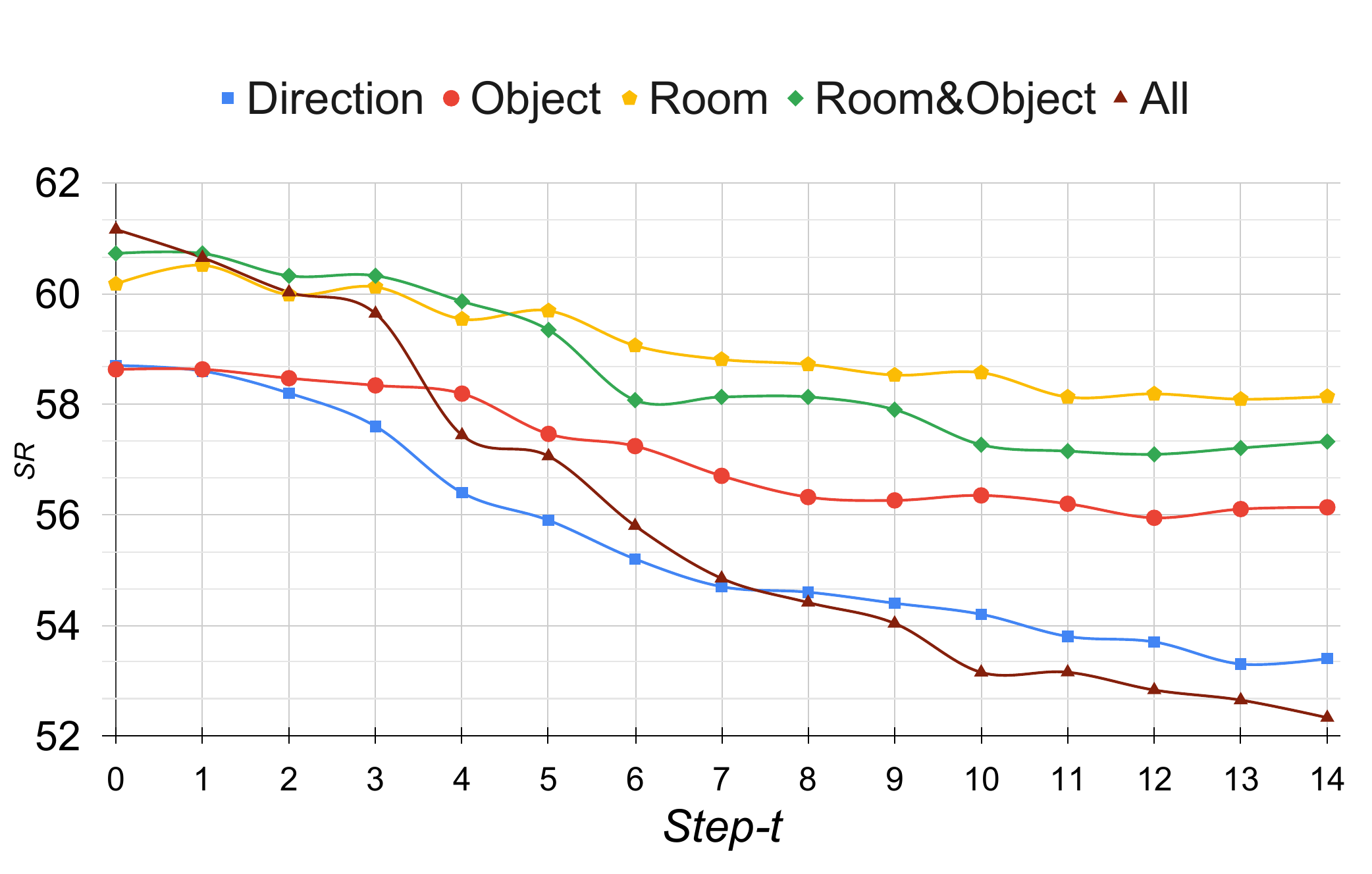}
    \caption{Success Rate upper bound ($\overline{\texttt{SR}}$) at different step-$t$.}
    \label{fig:sr_up_bound}
\end{figure}

\textbf{Can agents recover from instruction errors?}
Here, we want to show the capability of the agent to recover from different instruction errors at different steps. Thus, we want to simulate the following interaction from the human: \textit{``Sorry, the instruction I gave you before is wrong. This is the correct one: $<$instruction$>$"} where $<$\textit{instruction}$>$ is the correct instruction.
Specifically, for every perturbed episode $i$ in  $\mathcal{E}_p$, we let the agent navigate with the perturbed instruction for $t$-steps before providing the correct instruction at step $t+1$. 
As done before, it is important to note that instructions for correct episodes $i$ in $\mathcal{E}_c$ remain unchanged.
We show the results in Fig.~\ref{fig:sr_up_bound}, where we plot the \texttt{SR} at different $t$-steps for each benchmark present in \datasetName. As we can see, the \textit{Direction} and \textit{All} benchmarks exhibit the most significant decrease in SR. Note that, \textit{Direction} and \textit{All} have up to one and three errors per instruction, respectively.
Particularly for the \textit{Direction} and \textit{All} benchmarks, early error detection is crucial since these errors have a strong impact on the \texttt{SR}.

\begin{figure}[t!]
    \includegraphics[width=1\linewidth]{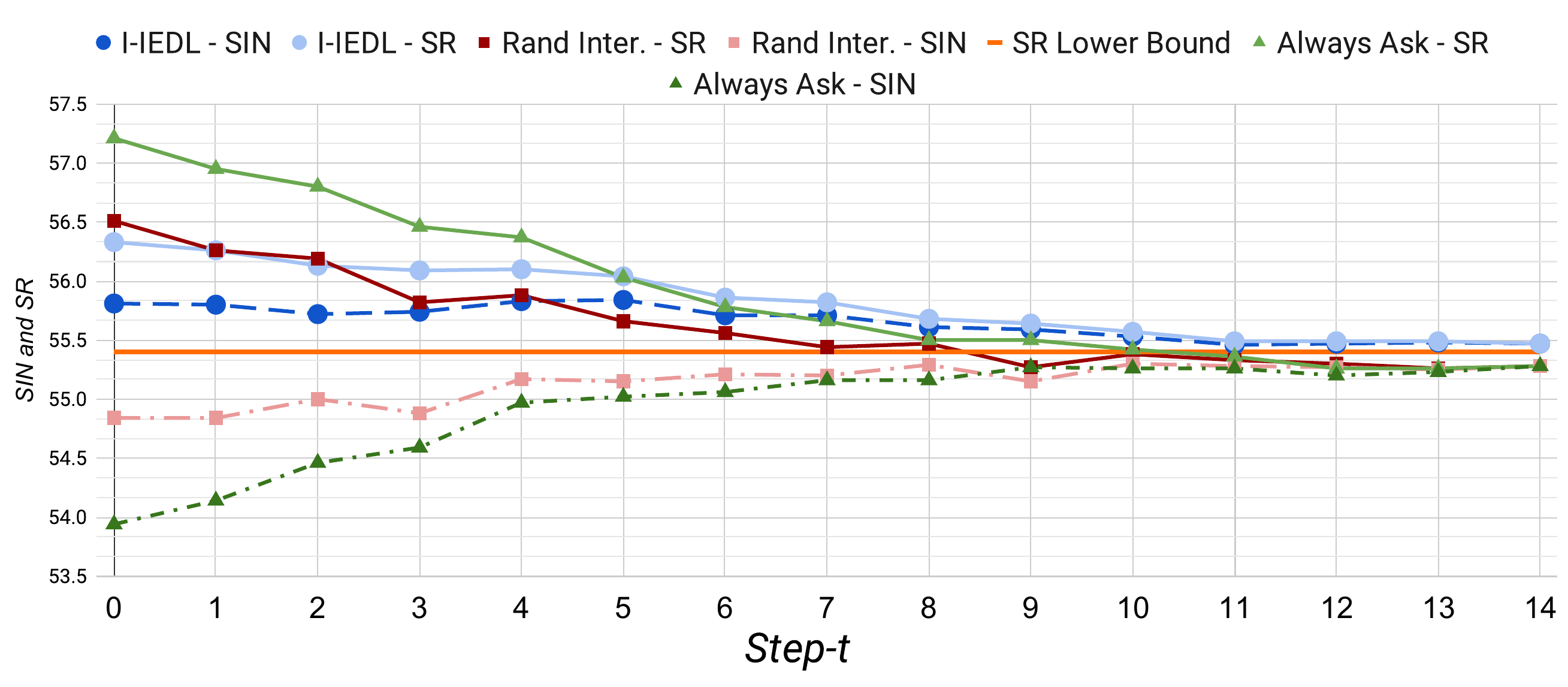}
    \caption{\texttt{SR} and \texttt{SIN} plotted at different step-$t$ for localization threshold $\tau_l=1$. Specifically, dashed lines indicate value for the \texttt{SIN} metric, while solid lines for \texttt{SR}. The \textit{``Always Ask"} baseline always interacts with the user from step-$t$ onwards.}
    \label{fig:sr_sin_different_steps}
\end{figure}

\textbf{How do SIN and SR evolve over steps?}
In this experiment, we want to analyze how \texttt{SR} and \texttt{SIN} evolve across steps using different interaction paradigms. In Fig.~\ref{fig:sr_sin_different_steps}, we thus show the \texttt{SR} values (solid lines) and \texttt{SIN} values (dashed lines) for different interaction paradigms called for different values step-$t$. Specifically, 
\textit{(i)}~\textit{``Random interaction"} and \textit{(ii)}~\textit{``Always Ask"}:~as described in Sec.~\ref{sec:exps}; and
\textit{(iii)}~\textit{I2EDL}:~as described in Sec.~\ref{sec:exps}, in which we set detection threshold $\tau_d=0.6$. For all the interaction paradigms, $\tau_l=1$.
As we can see from Fig~\ref{fig:sr_sin_different_steps}, the policy \textit{``Always Ask"} has the highest \texttt{SR}, since it continuously ask the user. On the other hand, this is very inconvenient, since humans do not want to be disturbed constantly. This behaviour is indeed reflected by our \texttt{SIN} metric, in which \textit{``Always Ask"} has the lowest \texttt{SIN} values. This behaviour is also reflected in the \textit{``Random interaction"}. Finally, we can see that our \textit{I2EDL} has the best compromise between \texttt{SR} and the number of interactions, behaviour that is correctly modelled by the \texttt{SIN} metric.

\textbf{Conclusions.} We presented a novel task, \taskAcronym, which enables interaction between an embodied agent and the human user to correct instruction errors while navigating to a goal described by textual navigation instructions. We proposed an effective baseline, \interactionLocalization, to perform error detection and localization in an online fashion. 
Compared to baselines, we showed that our proposed I2EDL is generally more effective in improving navigation performance when erroneous instructions are given, while lowering the interaction load.
Future works will investigate multi-modal interaction via images and text, with a thorough study on the user's cognitive load.

\bibliographystyle{IEEEtran}
\bibliography{IEEEabrv,references}

\begin{thebibliography}{10}
\providecommand{\url}[1]{#1}
\csname url@samestyle\endcsname
\providecommand{\newblock}{\relax}
\providecommand{\bibinfo}[2]{#2}
\providecommand{\BIBentrySTDinterwordspacing}{\spaceskip=0pt\relax}
\providecommand{\BIBentryALTinterwordstretchfactor}{4}
\providecommand{\BIBentryALTinterwordspacing}{\spaceskip=\fontdimen2\font plus
\BIBentryALTinterwordstretchfactor\fontdimen3\font minus \fontdimen4\font\relax}
\providecommand{\BIBforeignlanguage}[2]{{%
\expandafter\ifx\csname l@#1\endcsname\relax
\typeout{** WARNING: IEEEtran.bst: No hyphenation pattern has been}%
\typeout{** loaded for the language `#1'. Using the pattern for}%
\typeout{** the default language instead.}%
\else
\language=\csname l@#1\endcsname
\fi
#2}}
\providecommand{\BIBdecl}{\relax}
\BIBdecl

\bibitem{Taioli_2023_ICCV}
\BIBentryALTinterwordspacing
F.~Taioli, F.~Cunico, F.~Girella, R.~Bologna, A.~Farinelli, and M.~Cristani, ``{Language-Enhanced RNR-Map: Querying Renderable Neural Radiance Field Maps with Natural Language},'' in \emph{ICCVW}, Oct. 2023. [Online]. Available: \url{http://dx.doi.org/10.1109/iccvw60793.2023.00504}
\BIBentrySTDinterwordspacing

\bibitem{Anderson_2018_vln}
\BIBentryALTinterwordspacing
P.~Anderson, Q.~Wu, D.~Teney, J.~Bruce, M.~Johnson, N.~Sunderhauf, I.~Reid, S.~Gould, and A.~van~den Hengel, ``{Vision-and-Language Navigation: Interpreting Visually-Grounded Navigation Instructions in Real Environments},'' in \emph{CVPR}, Jun. 2018. [Online]. Available: \url{http://dx.doi.org/10.1109/cvpr.2018.00387}
\BIBentrySTDinterwordspacing

\bibitem{krantz_vlnce_2020}
\BIBentryALTinterwordspacing
J.~Krantz, E.~Wijmans, A.~Majumdar, D.~Batra, and S.~Lee, \emph{{Beyond the Nav-Graph: Vision-and-Language Navigation in Continuous Environments}}.\hskip 1em plus 0.5em minus 0.4em\relax ECCV, 2020, p. 104–120. [Online]. Available: \url{http://dx.doi.org/10.1007/978-3-030-58604-1_7}
\BIBentrySTDinterwordspacing

\bibitem{lloyd1989cognitive}
R.~Lloyd, ``Cognitive maps: Encoding and decoding information,'' \emph{AAG}, vol.~79, no.~1, 1989.

\bibitem{liben2022conceptual}
L.~S. Liben, ``Conceptual issues in the development of spatial cognition,'' in \emph{Spatial cognition}.\hskip 1em plus 0.5em minus 0.4em\relax Psychology Press, 2022, pp. 167--194.

\bibitem{couclelis1996verbal}
H.~Couclelis, ``{Verbal directions for way-finding: Space, cognition, and language},'' in \emph{The construction of cognitive maps}.\hskip 1em plus 0.5em minus 0.4em\relax Springer, 1996.

\bibitem{iedl}
\BIBentryALTinterwordspacing
F.~Taioli, S.~Rosa, A.~Castellini, L.~Natale, A.~D. Bue, A.~Farinelli, M.~Cristani, and Y.~Wang, ``{Mind the Error! Detection and Localization of Instruction Errors in Vision-and-Language Navigation},'' 2024. [Online]. Available: \url{https://arxiv.org/abs/2403.10700}
\BIBentrySTDinterwordspacing

\bibitem{matterport_dataset}
\BIBentryALTinterwordspacing
A.~Chang, A.~Dai, T.~Funkhouser, M.~Halber, M.~Niebner, M.~Savva, S.~Song, A.~Zeng, and Y.~Zhang, ``{Matterport3D: Learning from RGB-D Data in Indoor Environments},'' in \emph{3DV}, oct 2017, pp. 667--676. [Online]. Available: \url{https://doi.ieeecomputersociety.org/10.1109/3DV.2017.00081}
\BIBentrySTDinterwordspacing

\bibitem{habitat_19_iccv}
\BIBentryALTinterwordspacing
M.~Savva, A.~Kadian, O.~Maksymets, Y.~Zhao, E.~Wijmans, B.~Jain, J.~Straub, J.~Liu, V.~Koltun, J.~Malik, D.~Parikh, and D.~Batra, ``{Habitat: A Platform for Embodied AI Research},'' in \emph{ICCV}, Oct. 2019. [Online]. Available: \url{http://dx.doi.org/10.1109/iccv.2019.00943}
\BIBentrySTDinterwordspacing

\bibitem{hamt}
\BIBentryALTinterwordspacing
S.~Chen, P.-L. Guhur, C.~Schmid, and I.~Laptev, ``{History Aware Multimodal Transformer for Vision-and-Language Navigation},'' in \emph{NeurIPS}, vol.~34, 2021, pp. 5834--5847. [Online]. Available: \url{https://proceedings.neurips.cc/paper_files/paper/2021/file/2e5c2cb8d13e8fba78d95211440ba326-Paper.pdf}
\BIBentrySTDinterwordspacing

\bibitem{discrete_to_cont}
\BIBentryALTinterwordspacing
Y.~Hong, Z.~Wang, Q.~Wu, and S.~Gould, ``{Bridging the Gap Between Learning in Discrete and Continuous Environments for Vision-and-Language Navigation},'' in \emph{CVPR}, Jun. 2022. [Online]. Available: \url{http://dx.doi.org/10.1109/cvpr52688.2022.01500}
\BIBentrySTDinterwordspacing

\bibitem{an2023etpnav}
\BIBentryALTinterwordspacing
D.~An, H.~Wang, W.~Wang, Z.~Wang, Y.~Huang, K.~He, and L.~Wang, ``{ETPNav: Evolving Topological Planning for Vision-Language Navigation in Continuous Environments},'' \emph{arXiv arXiv:2304.03047}, 2023. [Online]. Available: \url{https://doi.org/10.48550/arXiv.2304.03047}
\BIBentrySTDinterwordspacing

\bibitem{an2023bevbert}
\BIBentryALTinterwordspacing
D.~An, Y.~Qi, Y.~Li, Y.~Huang, L.~Wang, T.~Tan, and J.~Shao, ``{BEVBert: Multimodal Map Pre-training for Language-guided Navigation},'' \emph{ICCV}, 2023. [Online]. Available: \url{https://doi.org/10.48550/arXiv.2212.04385}
\BIBentrySTDinterwordspacing

\bibitem{zhu2022diagnosing}
\BIBentryALTinterwordspacing
W.~Zhu, Y.~Qi, P.~Narayana, K.~Sone, S.~Basu, X.~Wang, Q.~Wu, M.~Eckstein, and W.~Y. Wang, ``{{Diagnosing Vision-and-Language Navigation: What Really Matters}},'' in \emph{NAACL}, Seattle, United States, Jul. 2022, pp. 5981--5993. [Online]. Available: \url{https://aclanthology.org/2022.naacl-main.438}
\BIBentrySTDinterwordspacing

\bibitem{limitation_vln_agent_aamas}
\BIBentryALTinterwordspacing
M.~Hahn, A.~Raj, and J.~M. Rehg, ``{Which way is ‘right’?: Uncovering limitations of Vision-and-Language Navigation models},'' in \emph{AAMAS}, 2023. [Online]. Available: \url{https://api.semanticscholar.org/CorpusID:253381693}
\BIBentrySTDinterwordspacing

\bibitem{diagnosin_env_bias}
\BIBentryALTinterwordspacing
Y.~Zhang, H.~Tan, and M.~Bansal, ``{Diagnosing the Environment Bias in Vision-and-Language Navigation},'' in \emph{IJCAI-PRICAI}, Jul. 2020. [Online]. Available: \url{http://dx.doi.org/10.24963/ijcai.2020/124}
\BIBentrySTDinterwordspacing

\bibitem{cvdn}
\BIBentryALTinterwordspacing
J.~Thomason, M.~Murray, M.~Cakmak, and L.~Zettlemoyer, ``{Vision-and-Dialog Navigation},'' in \emph{CoRL}, ser. Proceedings of Machine Learning Research, vol. 100.\hskip 1em plus 0.5em minus 0.4em\relax PMLR, 30 Oct--01 Nov 2020, pp. 394--406. [Online]. Available: \url{https://proceedings.mlr.press/v100/thomason20a.html}
\BIBentrySTDinterwordspacing

\bibitem{roman2020rmm}
\BIBentryALTinterwordspacing
H.~Roman~Roman, Y.~Bisk, J.~Thomason, A.~Celikyilmaz, and J.~Gao, ``{RMM: A Recursive Mental Model for Dialogue Navigation},'' in \emph{EMNLP}.\hskip 1em plus 0.5em minus 0.4em\relax Association for Computational Linguistics, 2020. [Online]. Available: \url{http://dx.doi.org/10.18653/v1/2020.findings-emnlp.157}
\BIBentrySTDinterwordspacing

\bibitem{banerjee2021robotslang}
\BIBentryALTinterwordspacing
S.~Banerjee, J.~Thomason, and J.~Corso, ``{The RobotSlang Benchmark: Dialog-guided Robot Localization and Navigation},'' in \emph{CoRL}, ser. Proceedings of Machine Learning Research, J.~Kober, F.~Ramos, and C.~Tomlin, Eds., vol. 155.\hskip 1em plus 0.5em minus 0.4em\relax PMLR, 16--18 Nov 2021, pp. 1384--1393. [Online]. Available: \url{https://proceedings.mlr.press/v155/banerjee21a.html}
\BIBentrySTDinterwordspacing

\bibitem{shrivastava2021visitron}
\BIBentryALTinterwordspacing
A.~Shrivastava, K.~Gopalakrishnan, Y.~Liu, R.~Piramuthu, G.~Tur, D.~Parikh, and D.~Hakkani-Tur, ``{VISITRON: Visual Semantics-Aligned Interactively Trained Object-Navigator},'' 2022. [Online]. Available: \url{http://dx.doi.org/10.18653/v1/2022.findings-acl.157}
\BIBentrySTDinterwordspacing

\bibitem{zhu2021self}
\BIBentryALTinterwordspacing
Y.~Zhu, Y.~Weng, F.~Zhu, X.~Liang, Q.~Ye, Y.~Lu, and J.~Jiao, ``{Self-Motivated Communication Agent for Real-World Vision-Dialog Navigation},'' in \emph{ICCV}, Oct. 2021. [Online]. Available: \url{http://dx.doi.org/10.1109/iccv48922.2021.00162}
\BIBentrySTDinterwordspacing

\bibitem{just_ask}
\BIBentryALTinterwordspacing
T.-C. Chi, M.~Shen, M.~Eric, S.~Kim, and D.~Hakkani-tur, ``{Just Ask: An Interactive Learning Framework for Vision and Language Navigation},'' vol.~34, no.~03, p. 2459–2466, Apr. 2020. [Online]. Available: \url{http://dx.doi.org/10.1609/aaai.v34i03.5627}
\BIBentrySTDinterwordspacing

\bibitem{vision_based_nav}
\BIBentryALTinterwordspacing
K.~Nguyen, D.~Dey, and B.~Brockett, Chrnd~Dolan, ``{Vision-Based Navigation With Language-Based Assistance via Imitation Learning With Indirect Intervention},'' in \emph{CVPR}, Jun. 2019. [Online]. Available: \url{http://dx.doi.org/10.1109/cvpr.2019.01281}
\BIBentrySTDinterwordspacing

\bibitem{avlen}
\BIBentryALTinterwordspacing
S.~Paul, A.~K. Roy-Chowdhury, and A.~Cherian, ``{AVLEN: Audio-Visual-Language Embodied Navigation in 3D Environments},'' 2022. [Online]. Available: \url{https://doi.org/10.48550/arXiv.2210.07940}
\BIBentrySTDinterwordspacing

\bibitem{help_anna}
\BIBentryALTinterwordspacing
K.~Nguyen and H.~Daumé~III, ``{Help, Anna! Visual Navigation with Natural Multimodal Assistance via Retrospective Curiosity-Encouraging Imitation Learning},'' in \emph{EMNLP-IJCNLP}, 2019. [Online]. Available: \url{http://dx.doi.org/10.18653/v1/d19-1063}
\BIBentrySTDinterwordspacing

\bibitem{anderson2018evaluation}
\BIBentryALTinterwordspacing
P.~Anderson, A.~Chang, D.~S. Chaplot, A.~Dosovitskiy, S.~Gupta, V.~Koltun, J.~Kosecka, J.~Malik, R.~Mottaghi, M.~Savva \emph{et~al.}, ``{On evaluation of embodied navigation agents},'' \emph{arXiv preprint arXiv:1807.06757}, 2018. [Online]. Available: \url{https://doi.org/10.48550/arXiv.1807.06757}
\BIBentrySTDinterwordspacing

\end{thebibliography}

\end{document}